\documentclass[review]{elsarticle}
\usepackage{lineno,hyperref}
\usepackage{graphicx}
\usepackage{pdfpages}
\usepackage{times}
\usepackage{epsfig}
\usepackage{amsmath,amssymb,amstext}
\usepackage{amssymb}
\usepackage{algorithm}
\usepackage{algorithmic}

\usepackage{booktabs}
\usepackage{epstopdf}
\usepackage{subfigure}
\usepackage{booktabs}
\usepackage{diagbox}
\usepackage{float}
\usepackage{makecell}
\usepackage{tabularx}

\usepackage{mathrsfs} 
\usepackage{hyperref}
\usepackage{lipsum}
\usepackage{amssymb}
\usepackage{amsmath}
\usepackage{mathrsfs} 
\usepackage{amsthm}
\usepackage{bbm}
\usepackage{booktabs}
\usepackage{threeparttable}
\usepackage{amsfonts}
\usepackage{multirow}
\usepackage{stmaryrd}
\modulolinenumbers[1]
\journal{Neurocomputing}
\begin{document}

\begin{frontmatter}
\title{Tucker decomposition-based Temporal Knowledge Graph Completion}

\author{Pengpeng Shao$^1$, Guohua Yang$^1$, Dawei Zhang$^1$, Jianhua Tao$^{1,2}$, Feihu Che$^1$, Tong Liu$^1$}

\address{$^1$National Laboratory of Pattern Recognition, Institute of Automation, Chinese Academy of Sciences; $^2$School of Artificial Intelligence, University of Chinese Academy of Sciences\\}

\begin{abstract}
Knowledge graphs have been demonstrated to be an effective tool for numerous intelligent applications. However, a large amount of valuable knowledge still exists implicitly in the knowledge graphs. To enrich the existing knowledge graphs, recent years witness that many algorithms for link prediction and knowledge graphs embedding have been designed to infer new facts. But most of these studies focus on the static knowledge graphs and ignore the temporal information that reflects the validity of knowledge. Developing the model for temporal knowledge graphs completion is an increasingly important task. In this paper, we build a new tensor decomposition model for temporal knowledge graphs completion inspired by the Tucker decomposition of order 4 tensor. We demonstrate that the proposed model is fully expressive and report state-of-the-art results for several public benchmarks. Additionally, we present several regularization schemes to improve the strategy and study their impact on the proposed model. Experimental studies on three temporal datasets (i.e. ICEWS2014,  ICEWS2005-15, GDELT) justify our design and demonstrate that our model outperforms baselines with an explicit margin on link prediction task.
\end{abstract}

\begin{keyword}
Temporal Knowledge Graphs \sep Tucker Tensor Decomposition \sep Link Prediction
\end{keyword}

\end{frontmatter}


\section{Introduction}
\label{sec::introduction}

Knowledge graphs (KGs) are graph-structured representations of knowledge and facts in the real-world, which are represented in the form of the triple (subject, predicate, object). KGs have been demonstrated to be available for various downstream tasks, such as recommendation~\cite{RecommenderSystems}, question answering~\cite{QA}, and information retrieval~\cite{InformationRetrieval}. 
However, a great deal of knowledge remains hidden in the KGs, that is, many links are missing between entities in KGs. In recent years, many algorithms for KGs completion are proposed to enrich KGs. As a representative method, KGs embedding is a current research hotspot and presents its efficiency and effectiveness on KGs completion. This type of method aims to model the nodes and relations in KGs and learn their low-dimensional hidden representation on the premise of preserving the graph structure and knowledge, then feeding the representation of each triple to score function for its validity.

It's worth noting that there are two kinds of knowledge in KGs, one is static knowledge and the other is temporal knowledge. The validity of static knowledge does not change over time, e.g., (WuHan, CityOf, China). While the temporal knowledge may not be universally true. In other words, the temporal knowledge is  true only in a specific period or a certain point in time. For instance, (Barack Obama, PresidentOf, America) is true only from 2008 to 2016, and (Einstein, WonPrize, Nobel Prize) only hold in 1922. However, most current works focus on the static KGs with an intuitive assumption that each triple in KGs is universally true and ignores the temporal scopes that reflect the validity of knowledge. 
In this work, we study the link prediction task~\cite{Linkprediction} on the temporal KGs which is also known as a standard completion task. The facts in temporal KGs are presented as (subject, predicate, object, timestamps), and the task is to answer the queries (subject, predicate, ?, timestamps) and (?, predicate, object, timestamps).

Recently, tensor factorization methods have been successfully applied to static KGs completion.~\cite{CanonicalTD, Tucker} apply Canonical Polyadic (CP) decomposition and Tucker decomposition to KGs completion, respectively. And they frame KGs completion problem as an order 3 tensor completion problem. Inspired by the CP decomposition of order 4 tensor, TNTComplex~\cite{TNTComplex} expresses temporal KGs completion problem as an order 4 tensor completion problem and presents an extension of ComplEx for temporal KGs completion. In essence, CP decomposition is a special case of Tucker decomposition, and the Tucker decomposition with a core tensor has a more powerful expressive ability than CP decomposition. Furthermore, it is an extremely hard problem to calculate the rank of tensors for CP decomposition, thus TNTComplex has to put great effort into selecting a proper rank of tensors manually for temporal KGs completion.

In this work, we build a TuckERT decomposition model for temporal KGs completion inspired by the Tucker decomposition of order 4 tensor. Essentially, the work expects to build the relationship between temporal information timestamps and the triple (subject, predicate, object). Specifically, if the facts (subject, predicate, object) hold, the temporal information underpinning the validity of the facts is hidden in entities or predicate. Thus the temporal information can be extracted from the entities or predicate of the triple. Conversely, given the 4-tuple facts, by associating the time-wise information with entities or predicate, the temporal fact that originally expressed by 4-tuple is now expressed implicitly by triple. Then we can perform tensor completion to achieve the task of temporal KGs completion. Similarly to TNTComplex~\cite{TNTComplex}, and De-simple~\cite{DESimplE}, we take the temporal and non-temporal knowledge into our decomposition model to handle the heterogeneity of knowledge. 
Our contributions are as follows:
\begin{itemize}
\item Developing a new tensor decomposition model for temporal KGs completion inspired by the Tucker decomposition of order 4 tensor.
\item Presenting several regularization schemes to improve our model and studying the impact of these regularizations on the proposed model.
\item Experimental studies on three temporal datasets demonstrate that our algorithm achieves state-of-the-art performance.
\end{itemize}

\section{Related Work} 

In this section, we introduce the previously proposed methods for static and temporal KGs completion, and most of the methods are based on tensor decomposition.
\subsection{Static KGs Completion} 
Static KGs completion methods can be broadly classified into three paradigms: Translational distance-based models such as TransE~\cite{TransE} and TransD~\cite{TransD}, Tensor factorization-based methods, and Neural network-based models including ConvE~\cite{ConvE} and R-GCN~\cite{R-GCN}. In particular, tensor decomposition has been favored in KGs completion for its high efficiency and powerful function. RESCAL~\cite{RESCAL} is the seminal tensor decomposition method for static KG completion, which imposes a score function of a bilinear product on the two entity vectors and predicate matrix. Although the model is fully expressive, it tends to arise over-fitting problems as the predicate matrix includes a large number of parameters. Latter, DisMult~\cite{DisMult} is aware of the above defects, then simplifying RESCAL by diagonalizing the predicate matrix. Which also poses a problem that the diagonal predicate matrix only models symmetric relation but asymmetric relation. To address this problem, ComplEx~\cite{ComplEx} projects entity and predicate embeddings into complex space to better model asymmetric relation. From another perspective, HoLE~\cite{HoLE} applies circular correlation operation to subject and object entity vectors to obtain a compositional vector, which then matches the predicate vectors to score the fact, thus the model absorbs the advantages of RESCAL and DistMult. The above methods based on CP tensor decomposition learn the subject and object entity representation independently. This is also the main reason for conducting link prediction poorly. In view of this, SimplE~\cite{SimplE} presents a new CP method that takes advantage of the inverse of the predicate to address the obstacle. In addition to the methods based on CP decomposition, TuckER~\cite{Tucker} based on Tucker decomposition also frames KGs completion as a 3rd-order binary tensor completion problem and factorizes the binary tensor of known facts into core tensor and three orthogonal matrixes.
\subsection{Temporal KGs Completion}
Most static completion models fail to take temporal dimension while learning embeddings of the KG elements. The temporal KGs completion remains a valuable but rarely studied research issue, and recent years witness that only a handful of temporal KGs completion models are presented.  

{\bf t-TransE}~\cite{tTransE}: To model the transformation between the time-aware predicate of two adjacent facts, t-TransE imposes temporal order constraints on the geometric structure of the embedding space to enforce the embeddings to be temporally consistent and more accurate. Then t-TransE optimizes the joint model consisting of temporal order constraints and the TransE model to make the embedding space compatible with the observed triple in the fact dimension.

{\bf HyTE}~\cite{HyTE}: Inspired by TransH~\cite{TransH}, HyTE associates temporal information with entity and predicate by projecting them to the hyperplane modeled by temporal information. Then HyTE accomplishes the embedding learning of entity and predicate which incorporate temporal information by minimizing translation distance.

{\bf TA-DistMult}~\cite{TADistMult}: To incorporate temporal information, TA-DistMult employs a recurrent neural network to learn the time-aware representation of predicate which then be utilized in DistMult and TransE. 

{\bf ConT}~\cite{ConT}: To model the cognitive function, the work generalized several static KGs approaches including Tucker and RESCAL to temporal/episodic  KGs. Tree and ConT are two novel generalizations of RESCAL to episodic tensors, and ConT obtains outstanding performance overall through introducing the latent representation of time for sparse episodic tensor.

{\bf DE-SimplE}~\cite{DESimplE}: Motivated by diachronic word embeddings, DE-SimplE, combining the diachronic entity embedding function with the static model SimplE, is capable of instructing the model to learn the temporal features of the entity at any point in time for temporal KG completion.

{\bf TNTComplEx}~\cite{TNTComplex}: Inspired by the canonical decomposition of order-4 tensor, TNTComplEx introduces an extension of ComplEx for temporal KG completion. Although TNTComplEx obtains considerable performance, it is hard to determine the rank of the tensor accurately.

All the above temporal methods associate time-wise information with entity and/or predicate to obtain temporal embedding, then equipping the embedding with static models for temporal KG completion. In line with that, we propose a Tucker decomposition of an order 4 tensor for temporal KG completion and introduce several kinds of regularization constraints to coordinate the proposed decomposition from different aspects. Experimental results show that our algorithm achieves state-of-the-art performance.

\section{Background}
CP decomposition~\cite{CP} factorizes a tensor into sum of component rank-one tensors. While Tucker decomposition~\cite{Tucker} is a form of high-order PCA, which factorizes a tensor into a core tensor multiplied by a matrix along each mode. Specifically, given a tensor $\mathcal{X} \in \mathbb{R}^{n_{1} \times n_{2} \times n_{3}}$, Tucker decomposition can factorize $\mathcal{X}$ along three mode into core tensor $\mathcal{G}$ and three matrix $\mathnormal{\bf A}, \mathnormal{\bf B}, \mathnormal{\bf C}$:
\begin{equation}
\begin{aligned}
\label{eq::Tucker_decomposition}
&\mathcal{X} \approx \mathcal{G} \times_{1} \mathnormal{\bf A} \times_{2} \mathnormal{\bf B}  \times_{3} \mathnormal{\bf C}
= \underset{p}\sum \underset{k}\sum \underset{q}\sum g^{(pkq)} a^{(p)}\circ b^{(k)}\circ c^{(q)}
= \llbracket{\mathcal{G}; \mathnormal{\bf A}, \mathnormal{\bf B}, \mathnormal{\bf C}}\rrbracket
\end{aligned}
\end{equation}
Where $\circ$ represents the vector outer product. The core tensor $\mathcal{G} \in \mathbb{R}^{r_{1} \times r_{2} \times r_{3}}$ can capture the information of interaction between the different components. The matrix in each mode $\mathnormal{\bf A} \in \mathbb{R}^{n_{1} \times r_{1}}$, $\mathnormal{\bf B} \in \mathbb{R}^{n_{2} \times r_{2}}$, $\mathnormal{\bf C} \in \mathbb{R}^{n_{3} \times r_{3}}$ are orthogonal to each other. $ \times_{n}$ suggests  the tensor product along the $\mathnormal{n}$-th mode. In addition,
if the core tensor is super-diagonal and satisfies $r_{1} = r_{2} = r_{3}$, the Tucker decomposition is equivalent to CP decomposition.

Latter, TuckER~\cite{Tucker} employs this type of decomposition for KGs completion, which views matrix $\mathnormal{\bf A}$ and  $\mathnormal{\bf C}$ as entity embedding $\bf E$, $\bf E = \bf A = \bf C$ $\in \mathbb{R}^{n_{e} \times d_{e}}$, with $n_{e}$ indicates the number of entities, $d_{e}$ suggests the dimensionality of entity embedding. Regarding $\bf B$ as predicate embedding $\bf R$, $\bf R$ = $\bf B$ $\in \mathbb{R}^{n_{r} \times d_{r}}$, where $n_{r}$ and $d_{r}$ denote the number of predicates and the dimensionality of predicate embedding, respectively. The scoring function is represented as follows:
\begin{equation}
\label{eq::TuckER}
\phi({e}_s, {e}_r, {e}_o) = \mathcal{W} \times_{1} \mathrm{e}_s \times_{2} \mathrm{e}_r  \times_{3} \mathrm{e}_o = \llbracket{\mathcal{W}; \mathrm{e}_s, \mathrm{e}_r, \mathrm{e}_o}\rrbracket 
\end{equation}
Where $\mathcal{W} \in \mathbb{R}^{d_{e} \times d_{r} \times d_{e}}$ also is core tensor, and the number of parameter in $\mathcal{W}$ only relys on the embedding dimensionality of entity and predicate, not on the number of entities or predicates. Meanwhile, TuckER also justifies that ComplEx based on CP decomposition is a special case of TuckER.
\section{The Proposed Model}

The temporal fact $(s, r, o, t)$ gives the triple $(s, r, o)$ a temporal label $t$ to ensure its accuracy. We argue that the temporal information is contained in the entity or predicate of the correct triple fact implicitly. For example, if the fact (Barack Obama, PresidentOf, America) is universally true, then the temporal element [2008, 2016] of the 4-tuple is contained implicitly in the subject entity of Barack Obama, or the predicate of PresidentOf, or the object entity of America. We can express the above three cases as the following forms:
\begin{itemize}
\item (Barack Obama $\&$ [2008, 2016], PresidentOf, America): Barack Obama of that period from 2008 to 2016 was President of the United States. That is, the entity of Barack Obama in the triple refers to Barack Obama of that period from 2008 to 2016.
\item (Barack Obama, PresidentOf $\&$ [2008, 2016], America): Barack Obama was President of that period from 2008 to 2016 of the United States.
\item (Barack Obama, PresidentOf, America $\&$ [2008, 2016]): Barack Obama was President of that United States of the period from 2008 to 2016. 
\end{itemize}
Although the timestamps information emphasizes different objects to express their temporal nature, respectively, these three facts accurately express one same meaning. Therefore, we can extract the temporal information $T$ from the triples to express the temporal facts in form of 4-tuple. Meanwhile, the embedding dimensionality of timestamps is the same as the embedding dimensionality of predicate or entity. Thus the TuckER decomposition can naturally be extended to the following form by adding temporal information $T$~\cite{TensorDA}:
\begin{equation}
\label{eq::TuckERT}
\phi(E, R, T) =\llbracket{\mathcal{M}; \mathnormal{\bf E}, \mathnormal{\bf R}, \mathnormal{\bf E}, \mathnormal{\bf T}}\rrbracket 
\end{equation} 
Where $\mathcal{M} \in \mathbb{R}^{d_{e} \times d_{r} \times d_{e} \times d_{t}}$ is core tensor. We call this kind of decomposition TuckERT, E represents entities, R denotes predicates, and T indicates temporal information.
Given the temporal fact $(\mathrm{e}_s, \mathrm{e}_r, \mathrm{e}_o,\mathrm{e}_t)$, the TuckERT decomposition is expressed in following inner product form:
\begin{equation}
\label{eq::TuckERT-1}
\phi(E, R, T)_{s,r,o,t}= \left \langle \mathcal{M}; \mathrm{e}_s, \mathrm{e}_r, \mathrm{e}_o,\mathrm{e}_t\right \rangle
\end{equation} 

As described in the above example, the time-wise information can be employed to associate with the subject, predicate or object to obtain time-dependent embedding equivalently. In addition, obtaining the time-dependent embedding can be viewed as the inverse process of TuckERT decomposition:
\begin{equation}
\begin{aligned}
\label{eq::TuckERT-2}
 \left \langle \mathcal{M}; \mathrm{e}_s, \mathrm{e}_r, \mathrm{e}_o,\mathrm{e}_t \right \rangle = 
\left \langle \mathcal{W}; \mathrm{e}_s \odot \mathrm{e}_t , \mathrm{e}_r, \mathrm{e}_o\right \rangle
= \left \langle \mathcal{W}; \mathrm{e}_s, \mathrm{e}_r \odot \mathrm{e}_t , \mathrm{e}_o\right \rangle 
= \left \langle \mathcal{W}; \mathrm{e}_s, \mathrm{e}_r, \mathrm{e}_o\odot \mathrm{e}_t \right \rangle 
\end{aligned}
\end{equation} 
Here, $\odot$ denotes dot product, $\mathcal{W} \in \mathbb{R}^{d_{e} \times d_{r} \times d_{e}}$ is the folding version of $\mathcal{M}$ on dimension 2 and 4. Then, we can conduct temporal KGs completion as static methods. In addition, it's worth noting that some facts may vary with time-wise dimension, while some facts are independent of time. As mentioned above, the fact (Barack Obama, PresidentOf, America) hold only from 2008 to 2016, (Einstein, WonPrize, Nobel Prize) is true only at 1922. While the correctness of the facts (WuHan, CityOf, China) does not change over time. To model the two kinds of knowledge, temporal facts and non-temporal facts, following~\cite{TNTComplex,DESimplE}, we propose a TuckERTNT model which is a variant of TuckERT: 
\begin{equation}
\label{eq::TuckERTNT}
\phi(E, R, T) =\llbracket{\mathcal{M}; \mathnormal{\bf E}, \mathnormal{\bf R}, \mathnormal{\bf E}, \mathnormal{\bf T}}\rrbracket + \llbracket{\mathcal{M}; \mathnormal{\bf E}, \mathnormal{\bf R}, \mathnormal{\bf E}, \mathrm{\bf 1}}\rrbracket
\end{equation}

Similarly, given the temporal fact $(\mathrm{e}_s, \mathrm{e}_r, \mathrm{e}_o,\mathrm{e}_t)$, the specific form of the TuckERTNT decomposition can be described as follows:
\begin{equation}
\label{eq::obtain E}
\phi(E, R, T)_{s,r,o,t}= \left \langle \mathcal{W}; \mathrm{e}_s,  \mathrm{e}_r^t \odot \mathrm{e}_t + \mathrm{e}_r, \mathrm{e}_o\right \rangle 
\end{equation}
Where NT is an abbreviation of non-temporal. Compared with TNTComplEx, our model does not select the rank of tensors manually. We associate the temporal information with the predicate to learn time-dependent embeddings on the basis of the Tucker decomposition and use the parameters of the core tensor to increase the level of interaction in each dimension between the entity and time-dependent predicate, thus obtaining state-of-the-art performance. Furthermore,~\cite{Tucker} proved that ComplEx is equivalent to TuckER on the premise of imposing certain constraints on the core tensor, or ComplEx is a special case of TuckER. We give a similar result that TComplEx can be viewed as equivalent to TuckERT on the premise of certain constraints. 

\renewcommand\arraystretch{1.3}
\begin{table}
\centering
\caption{Number of parameters of the proposed method and baseline models. $d$ represents the embedding dimensionality, $r$ denotes the rank of a tensor, and $2r = d$.}  
\label{tab:number_parameter} 
\begin{tabularx}{10cm}{llX}
\hline                     
De-SimplE  & $d((3\gamma + (1-\gamma))\lvert E \rvert + \lvert R \rvert)$  \\ 
TComplEx   & $2r(\lvert E \rvert + \lvert T \rvert + 2\lvert R \rvert)$  \\  
TNTComplEx   & $2r(\lvert E \rvert + \lvert T \rvert + 4\lvert R \rvert)$  \\  
\hline  
TuckERT   & $d(\lvert E \rvert + \lvert T \rvert + 2\lvert R \rvert) + d^3$\\  
TuckERTNT  & $d(\lvert E \rvert + \lvert T \rvert + 4\lvert R \rvert) + d^3$ \\  
\hline  
\end{tabularx}  
\end{table}

\subsection{Learning}
We employ data augmentation technique to add reciprocal predicates (object, predicate$^{-1}$, subject, timestamps) into training sets. The model parameters are learned utilizing stochastic gradient descent with mini-batches. Then we expect to minimize the instantaneous multi-class loss~\cite{CanonicalTD} to train our model :
\begin{equation}
\label{eq::X_decomposition}
\mathcal{L}(\theta) = - \phi(\theta; s, r, o, t) + \log\bigg(\underset{o^{'}}\sum \exp \Big(\phi(\theta; s, r, o', t)\Big) \bigg)
\end{equation}
Here, $(s, r, o, t)$ is a positive sample, $(s, r, o', t)$ represents the negative sample obtained by replacing the true object with a false.
Note that, the loss function can only be used to train the model to answer the queries of this form (subject, predicate, ?, timestamps). Due to the existence of inverse samples (object, predicate$^{-1}$, subject, timestamps), answering the queries of the type (object, predicate$^{-1}$, ?, timestamps) already includes answering (?, predicate, object, timestamps).

In addition, we expect that imposing regularization constraints on the matrix factorized from the tensor to incorporate the prior knowledge of temporal mode and avoid over-fitting. Referring to a prior assumption of the node of undirected graph that similar nodes have close low-dimensional representations, we expect the timestamps embedding satisfy the smooth constraint~\cite{TNTComplex} that neighbouring timestamps have close representations. The time-wise regularization is formulated as follows:
\begin{equation}
\label{eq::regularization_timestamps}
\mathcal{S}(T) = \frac{\lambda}{|T| -1}\underset{i=1}{\overset{|T| -1}\sum}{\|e_{t_{i+1}} - e_{t_{i}}\|_p^q}
\end{equation}

To prevent model from over-fitting, we impose embedding regularization constraints on subjects, temporal and non-temporal predicates, object, and core tensor. In addition, driven by the motivation that studying the impact of different embedding regularizations, we introduce the following two kinds of embedding regularization schemes. Due to core tensor contains large amount of parameters, we argue that the constraint on core tensor is important and may have big impact on our model. Therefore, on the basis of each type of regularization scheme, we study the impact of imposing constraints on the core tensor and without constraints, respectively. 
\begin{equation}
\begin{aligned}
\label{eq::regularization_entity_F}
&\mathcal{R}_{F}(E) = \frac{\alpha}{3}(2\|e_s\|^{k} _{F} + \|e_r^t \odot e_t\|^{k}_{F}+ 2\|e_o\|^{k}_{F}+ \|e_r\|^{k}_{F})\\
&\mathcal{R}_{F}( E, \mathcal{W}) =\frac{\alpha}{4}(2\|e_s\|^{k}_{F} + \|e_r^t \odot e_t\|^{k}_{F}+ 2\|e_o\|^{k}_{F}+ \|e_r\|^{k}_{F}
+ \|\mathcal{W}\|^{k}_{F})
\end{aligned}
\end{equation}
Where, $\|\bullet\|_{F}$ represents Frobenius norm, $\|\bullet\|^k$ is the $k$ power of tensor norm.
\begin{equation}
\label{eq::regularization_entity}
\begin{aligned}
&\mathcal{R}_{p}(E)= \frac{\alpha}{3}(2\|e_s\|_{p}^{q} + \|e_r^t \odot e_t\|_{p}^{q} + 2\|e_o\|_{p}^{q} + \|e_r\|_{p}^{q})\\
&\mathcal{R}_{p}(E, \mathcal{W}) =\frac{\alpha}{4}(2\|e_s\|_{p}^{q} + \|e_r^t \odot e_t\|_{p}^{q} + 2\|e_o\|_{p}^{q} + \|e_r\|_{p}^{q} 
+ \|\mathcal{W}\|_{p}^{q})\\
\end{aligned}
\end{equation}
Here,  $\|\bullet\|_{p}$ is the $l_p$ norm of the matrix, and $\|\bullet\|^q$ denotes the $q$ power of tensor norm. 
Considering instantaneous multi-class loss and the above two classes of regularization term jointly, we train our model through minimizing the following loss function:
\begin{equation}
\label{eq::X_decomposition2}
\mathcal{L} = \mathcal{L}(\theta) + \mathcal{S}(T) + \mathcal{R}_{*}(E, \cdot)
\end{equation}
Where $\mathcal{R}_{*}(E, \cdot)$ represents one of four embedding regularizations mentioned above.

\subsection{Time Complexity and Parameter Growth}
Table~\ref{tab:number_parameter} presents the numbers of parameters for the proposed models TuckERT, TuckERTNT, and baseline models De-SimplE, TComplEx and its variant TNTComplEx. The number of parameters in baseline models increases linearly with respect to the number of entities and predicates or embedding dimensionality $d$. While the number of parameters in our model grows three times with embedding dimensionality $d$ as the three-dimension core tensor depends only on the embedding dimensionality. Consequently, the time complexity for the proposed models TuckERT and TuckERTNT is $\mathcal{O}(d^3)$. As for De-SimplE, TComplEx, and TNTComplEx, they have a time complexity of $\mathcal{O}(d)$. It can be viewed that our model includes more parameters on the premise of the fixed number of entities, predicates, and embedding dimensionality. However, it has been argued that a model with many parameters tends to arise over-fitting and scalability problems, thus resulting in poor performance. Table~\ref{tab:performance_comparison} suggests that the proposed model can better fit the large-scale discrete temporal data by using more controllable parameters compared with the baseline models. Accordingly, an important challenge in designing a tensor decomposition model for temporal knowledge completion is the trade-off between model parameters and data, as well as the trade-off between expressiveness and model complexity. 

\subsection{Expressivity Analysis}
Full expressiveness is a very important property for the KG completion model, which refers to the ability of a model to correctly distinguish the positive facts from the negatives through learning. The proof of full expressiveness about our model is introduced as follows.
TuckERT is fully expressive for temporal knowledge graph completion.

Given a 4-tuple $(e_{s}, e_{r}, e_{o}, e_{t})$, where $e_{s}, e_{o} \in \mathbb{R}^{n_e \times d_e}$ are one-hot binary vector representations of subject and object, $e_{r} \in \mathbb{R}^{n_r \times d_r}$,  $e_{t} \in \mathbb{R}^{n_t \times d_r}$ are one-hot binary vector representations of predicate and timestamps, respectively. Here, the embedding dimensionality satisfies $d_e = n_e$, $d_r = n_r$. We set the $p$-th element of the binary vector $e_{s}$,  $k$-th element of  $e_{r}$,  $q$-th element of $e_{o}$, $r$-th element of  $e_{t}$  to be 1, all other elements to be 0. Moreover, we set the $pkqr$-th element of the tensor $\mathcal{M} \in \mathbb{R}^{d_{e} \times d_{r} \times d_{e}\times d_{r}}$ to 1 if the temporal fact $(e_{s}, e_{r}, e_{o}, e_{t})$ holds and -1 otherwise. According to TuckERT decomposition:

\begin{equation}
\begin{aligned}
\label{eq::TuckERTprove}
\phi(E, R, T)_{s,r,o,t}= \left \langle \mathcal{M}; \mathrm{e}_s, \mathrm{e}_r, \mathrm{e}_o,\mathrm{e}_t\right \rangle
= \underset{p}\sum \underset{k}\sum \underset{q}\sum \underset{r}\sum \mathcal{M}^{(pkqr)}\mathrm{e}_s^{(p)}\circ \mathrm{e}_r^{(k)}\circ \mathrm{e}_o^{(q)}\circ \mathrm{e}_t^{(r)}
\end{aligned} 
\end{equation} 
the inner product of the entity embeddings, the predicate embedding and time embedding with the core tensor is capable of representing the original temporal tensor accurately. And by modulating the parameters in core tensor, the model can completely distinguish the positive samples from the negative. 

From another perspective, we can regard the core tensor $\mathcal{M}$ as a linear classifier in high dimensional space, which possesses the ability to distinguish the positive and negative samples in low dimensional space through learning.  

\renewcommand\arraystretch{1} 
\begin{table*} 
\small
\caption{Dataset statistics.} 
\centering 
\label{tab:datasets} 
\begin{tabular}{l||l|l|l|l|l|l p{0.5cm}}  
DataSets & Entities & Predicates & Timestamps& Training & Validation & Test\\ \hline  
 \makecell[c]{ICEWS2014} &  \makecell[c]{7,128} &  \makecell[c]{230} & \makecell[c]{365} &  \makecell[c]{72,826} &  \makecell[c]{8,941} &  \makecell[c]{8,963}\\  
 \makecell[c]{ICEWS05-15} &  \makecell[c]{10,488} &  \makecell[c]{251} & \makecell[c]{4,017}  &  \makecell[c]{386,962} &  \makecell[c]{46,275} &  \makecell[c]{46,092}\\  
 \makecell[c]{GDELT} &  \makecell[c]{500} &  \makecell[c]{20} & \makecell[c]{366} &  \makecell[c]{2,735,685} &  \makecell[c]{341,961} &  \makecell[c]{341,961}\\  
\end{tabular} 
\end{table*}

\section{Experiment Result}

In this section, We introduce the experiment setup including parameters, datasets, and evaluation metrics, and Baselines. In addition, we also provide comparison results from different aspects with other popular trackers on different datasets.

\subsection{Experiment Setup}

\textbf{Parameters Settings:} We implement the proposed model and baseline models in Pytorch~\cite{pytorch} framework. The two variants of our model are optimized by Adgard algorithm~\cite{Adam} with batch size of 1000 and learning rate of 0.2. To evaluate the impact of embedding dimensionality on the performance of link prediction, we vary the embedding dimensionality in the range $\{32, 64, 100, 200, 300, 400\}$. Considering the efficiency and effectiveness of the model, the embedding dimensionality is eventually set to 300. Time weight coefficient $\lambda$  and embedding weight coefficient $\alpha$ are set as 0.01 and 0.002, respectively. The norm $p$ of the tensor is set to 4, the power $q$, and $k$ of the tensor norm are set to 2 and 1, respectively. For generality and fairness, we consistently apply the above parameters to perform our models on the three datasets. In addition, we reproduce the results of TNTComplEx on the GDELT dataset with the tensor rank of 256 for model comparison.

\textbf{Datasets:} We evaluate the proposed model by utilizing the following three standard benchmarks for temporal KGs completion. The details of these dataset statistics are presented in Table~\ref{tab:datasets}.

\begin{itemize}
\item ICEWS 2014: The ICEWS (Integrated Crisis Early Warning System)~\cite{ICEWS} dataset is the collection of 4-tuple which is extracted from digital and social news about political events. Intuitively, ICEWS 2014~\cite{TADistMult} sub-sampling from ICEWS is the temporal facts occurring in 2014, and it contains 7128 entities, 230 predicates, and 365 timestamps.
\item ICEWS 05-15: Similarly to ICEWS 2014, ICEWS 05-15~\cite{TADistMult} is another subset of ICEWS. This dataset corresponding to the temporal facts from 2005 to 2015 has 10488 entities, 251 predicates, and 4017 timestamps.  
\item GDELT: GDELT (Global Database of Events, Language, and Tone)~\cite{GDELT} is a repository that contains human social relationships. we implement our models and baselines on its subset dataset~\cite{subGDELT}, which corresponds to the facts from 2015 to 2016 and contains 500 entities, 20 predicates, and 366 timestamps. 
\end{itemize}

\textbf{Baselines :}  To evaluate the performance of our proposed TuckERT and TuckERTNT, we compare it with 8 state-of-the-art baseline methods introduced briefly in related work, including t-TransE, HyTE, TA-DistMult, ConT,  three variants of DE-SimplE, and TNTComplEx.

\subsection{Link Prediction}
Given the incomplete temporal data, the task of link prediction is to predict the missing entity. More specifically, this task answers the queries of the form  (subject, predicate, ?, timestamps) and  (?, predicate, object, timestamps). For the above two queries, we employ mean reciprocal rank (MRR) and Hit@$n$ to measure the prediction level of our model. MRR is the average of the reciprocal of the mean rank (MR) assigned to the true triple overall candidate triples, Hits@$n$ measures the percentage of test-set rankings where a true triple is ranked within the top $n$ candidate triples. 
While MRR is widely used as an evaluation indicator of inference in the literature as MRR is more stable~\cite{MRR} than MR which is highly susceptible to a bad prediction. We denote $k_{f, s}$ and $k_{f, o}$ as the ranking for subject $s$ and object $o$ for the two queries, respectively. MRR and Hit@$n$ are defined as follows. MRR: $\frac{1}{2 \ast \vert test\vert} \sum_{f = (s, r, o, t) \in test} (\frac{1}{k_{f, o}} + \frac{1}{k_{f, s}})$. Hit@$n$ :$\frac{1}{2 \ast \vert test\vert} \sum_{f = (s, r, o, t) \in test} (\mathbbm{1}_{k_{f, o}\leqslant n}+ \mathbbm{1}_{k_{f, s}\leqslant n})$, where $\mathbbm{1}_{variable}$ is 1 if $\mathnormal{variable}$ holds and 0 otherwise. 

\textbf{Results and Analysis}
\renewcommand{\arraystretch}{1.4} 
\begin{table*}[tp]
\LARGE
\centering
\caption{The results of Link prediction on ICEWS14, ICEWS05-15 and GDELT datasets. Best results are in bold.}
\label{tab:performance_comparison}
\resizebox{\textwidth}{40mm}{
\begin{tabular}{l|llll|llll|llll}
\toprule
\multirow{4}{*}{\huge Method}&
\multicolumn{4}{c|}{ ICEWS14}&\multicolumn{4}{c|}{ ICEWS05-15}&\multicolumn{4}{c}{ GDELT}\cr
\cmidrule(lr){2-5} \cmidrule(lr){6-9} \cmidrule(lr){10-13}
&MRR&Hit@1&Hit@3&Hit@10&MRR&Hit@1&Hit@3&Hit@10&MRR&Hit@1&Hit@3&Hit@10\cr
\midrule
t-TransE&0.255&0.074&\makecell[c]{-}&0.601&0.271&0.084&\makecell[c]{-}&0.616&0.115&0.0&0.160&0.318\cr
HyTE&0.297&0.108&0.416&0.655&0.316&0.116&0.445&0.681&0.118&0.0&0.165&0.326\cr
TA-DistMult&0.477&0.363&\makecell[c]{-}&0.686&0.474&0.346&\makecell[c]{-}&0.728&0.206&0.124&0.219&0.365\cr
ConT&0.185&0.117&0.205&0.315&0.163&0.105&0.189&0.272&0.144&0.080&0.156&0.265\cr
De-TransE&0.326&0.124&0.467&0.686&0.314&0.108&0.453&0.685&0.126&0.0&0.181&0.350\cr
De-DistMult&0.501&0.392&0.569&0.708&0.484&0.366&0.546&0.718&0.213&0.130&0.228&0.376\cr
De-simplE&0.526&0.418&0.592&0.725&0.513&0.392&0.578&0.748&0.230&0.141&0.248&0.403\cr
TNTComplEx&0.560&0.460&0.610&0.740&0.600&0.500&0.650&0.780&0.224&0.144&0.239&0.381\cr
\hline
TuckERT&0.594&0.518&0.640&0.731&0.627&0.550&0.674&0.769&{\bf0.411} &{\bf0.310} &{\bf0.453} &{\bf0.614}\cr
TuckERTNT&{\bf 0.604} &{\bf 0.521} &{\bf 0.655} &{\bf 0.753}&{\bf 0.638} &{\bf 0.559} &{\bf 0.686} &{\bf 0.783}&0.381&0.283&0.418&0.576\cr
\bottomrule
\end{tabular}}
\end{table*}

We evaluate two variants of our model on ICEWS14, ICEWS05-15, and GDELT datasets, respectively. The evaluation results against the baseline models  are presented in Table~\ref{tab:performance_comparison}, and the analysis details of the results are described as follows. From the comparative results, we  conclude that the proposed method TuckERT and TuckERTNT outperform DE-SimplE and TNTComplEx explicitly on the evaluation datasets. Especially in the GDELT dataset, the MRR performance of our model is almost twice of the baseline model TNTComplEx, which also suggests the superiority of the extension of Tucker decomposition in temporal KGs completion. Moreover, as presented in Table~\ref{tab:performance_comparison}, GDELT includes a large number of news facts in the global world for two years, this shows that the facts in GDELT are updated very quickly, thus GDELT has a stronger temporal property than ICEWS14 and ICEWS05-15. This may be the main reason why TuckERT obtains better performance than TuckERTNT that is affected by some non-temporal factors on the GDELT dataset. 

\subsection{Ablation Study}
To further evaluate the proposed model, we study the impact of regularization and embedding dimensionality on our model trained on the ICEWS14 dataset, respectively. In addition, we compare the stability of our model with baseline models.

\textbf{Impact of Regularization}
\renewcommand{\multirowsetup}{\centering} 
\begin{table}[tb]
\centering
\fontsize{9}{9}\selectfont
\caption{The performance of TuckERTNT model with different embedding dimensionalities trained on the ICEWS14 dataset.} 
\label{tab:impact_embedding_dimensionality} 
\begin{tabular}{l||l|l|l|lp{3cm}}
Dimensionality & MRR & Hit@1 & Hit@3 & Hit@10  \\ \hline
\makecell[c]{32} & 0.478 & 0.373 & 0.533 & 0.678 \\ 
\makecell[c]{64} & 0.517 & 0.429 & 0.566 & 0.682 \\    
\makecell[c]{100} & 0.554 & 0.475 & 0.602 & 0.698 \\  
\makecell[c]{200} & 0.594 & 0.513 & 0.642 & 0.740 \\  
\makecell[c]{300}& 0.604 & 0.521 & 0.655 & 0.753\\
\makecell[c]{400}& 0.608 & 0.524 & 0.661 & 0.761\\
\end{tabular}  
\end{table}  

As presented in Table~\ref{tab:impact_regularization}, on ICEWS2014 dataset, we study the impact of time regularization and different embedding regularizations on the TuckERTNT model. 
From the results, it can be concluded that, compared with embedding regularizations, the time-wise regularization contributes more to the performance and brings a performance gain of 2.4$\%$ in MRR. Without combining the time-wise regularization, the embedding regularization with Frobenius norm has a negative impact, while the embedding regularization with $l_{p}$ norm brings a performance gain of 1.0$\%$ in MRR. Meanwhile, we can also find that imposing constraints on the core tensor has a similar impact on our performance as without constraints. From another perspective, when considering the time-wise regularization jointly, the embedding regularization with Frobenius norm and $l_{p}$ norm bring performance gain of 0.5$\%$ and 0.8$\%$ in MRR, respectively. As described in~\cite{CanonicalTD}, embedding regularization does not dramatically help the CP tensor decomposition model. According to our results above, we find that the embedding regularization also does not contribute more to the Tucker decomposition model. Furthermore, the scheme of imposing constraints on the core tensor does not have much positive effect on our model.
\begin{table} 
\centering
\fontsize{8}{9}\selectfont
\caption{The impact of time-wise regularization and embedding regularization on TuckERTNT model trained on ICEWS14 dataset. Best results are in bold.} 
\label{tab:impact_regularization}
\begin{tabular}{l||l|l|l|lp{3.5cm}}  
TuckERTNT  Loss& MRR & Hit@1 & Hit@3 & Hit@10  \\ \hline  
$\mathcal{L}(\theta)$& 0.571 & 0.492 & 0.614 & 0.722 \\     
$\mathcal{L}(\theta) + \mathcal{R}_{F}(E)$& 0.565 & 0.488 & 0.606 &0.714\\   
$\mathcal{L}(\theta) + \mathcal{R}_{F}(E, \mathcal{W})$& 0.565 & 0.483 & 0.608 & 0.720\\
$\mathcal{L}(\theta) + \mathcal{R}(E)$ & 0.581 & 0.501 & 0.626 & 0.729\\ 
$\mathcal{L}(\theta) + \mathcal{R}(E, \mathcal{W})$ & 0.581 & 0.498 & 0.629 & 0.734\\ 
$\mathcal{L}(\theta) + \mathcal{S}(T) $ & 0.596 & 0.515 & 0.650 & 0.738 \\
\hline      
$\mathcal{L}(\theta)+ \mathcal{S}(T)+ \mathcal{R}_{F}(E) $ & 0.598 & 0.518 & 0.649 & 0.742\\
$\mathcal{L}(\theta)+ \mathcal{S}(T)+ \mathcal{R}_{F}(E, \mathcal{W}) $ & 0.601 & 0.519 & 0.650 & 0.749\\
$\mathcal{L}(\theta)+ \mathcal{S}(T)+ \mathcal{R}(E) $ & 0.602 & 0.520 & 0.651 & 0.747\\
$\mathcal{L}(\theta)+ \mathcal{S}(T)+ \mathcal{R}(E,\mathcal{W}) $ & {\bf0.604} & {\bf0.521} & {\bf0.655} & {\bf0.753}\\
\end{tabular}  
\end{table}
\begin{figure}[http]  
\centering
\includegraphics[height=6cm, width=7.5cm]{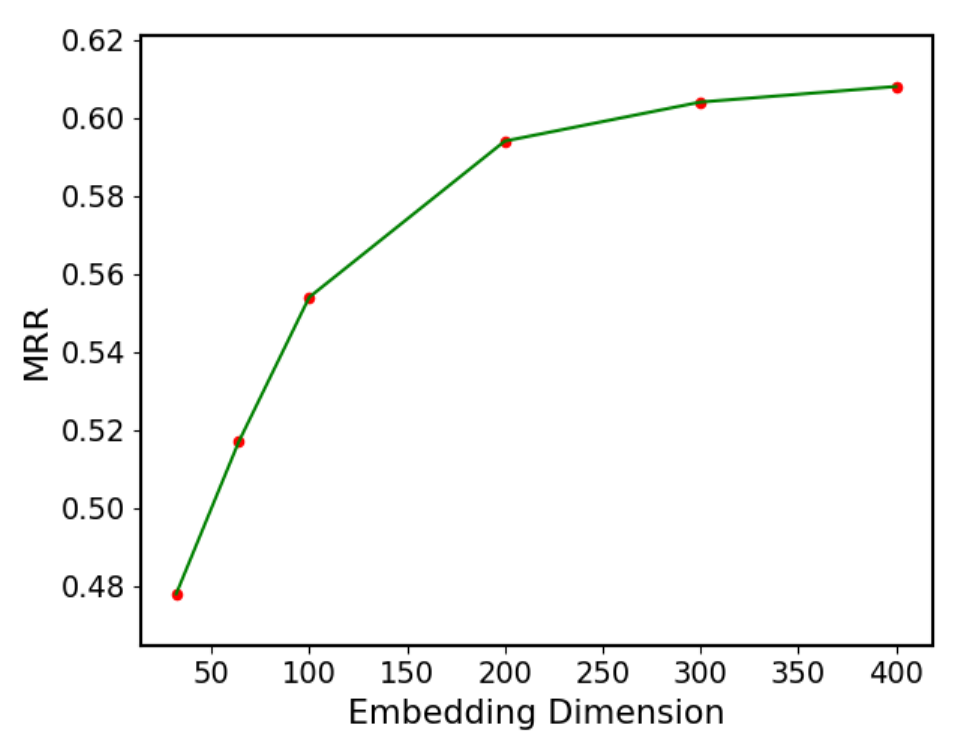} 
\caption{The MRR performance of TuckERTNT model with different embedding dimensionalities trained on the ICEWS14 dataset.} 
\label{fig::MRR_impact_embedding_dimension}
\end{figure}

\textbf{Impact of  Embedding Dimensionality}	

 To evaluate the impact of embedding dimensionality on our model, we vary the embedding dimensionality of the entity and predicate from 32 to 400 and report MRR and Hits@$n$ results of the TuckERTNT model on the ICEWS14 dataset in Table~\ref{tab:impact_embedding_dimensionality}. Figure~\ref{fig::MRR_impact_embedding_dimension} suggests the MRR performance trend with respect to embedding dimensionality. From the results, we can conclude that the MRR performance increases with embedding dimensionality growth until reaching a relatively steady state. While the higher embedding dimensionality means that the core tensor contains more parameters, thus leading to a sharp drop in efficiency while improving performance. Therefore, considering the efficiency and effectiveness of the proposed model jointly, the embedding dimensionality is eventually set to 300.

\textbf{Training Curve}
 
\begin{figure}[http]  
\centering
\includegraphics[height=6cm, width=7cm]{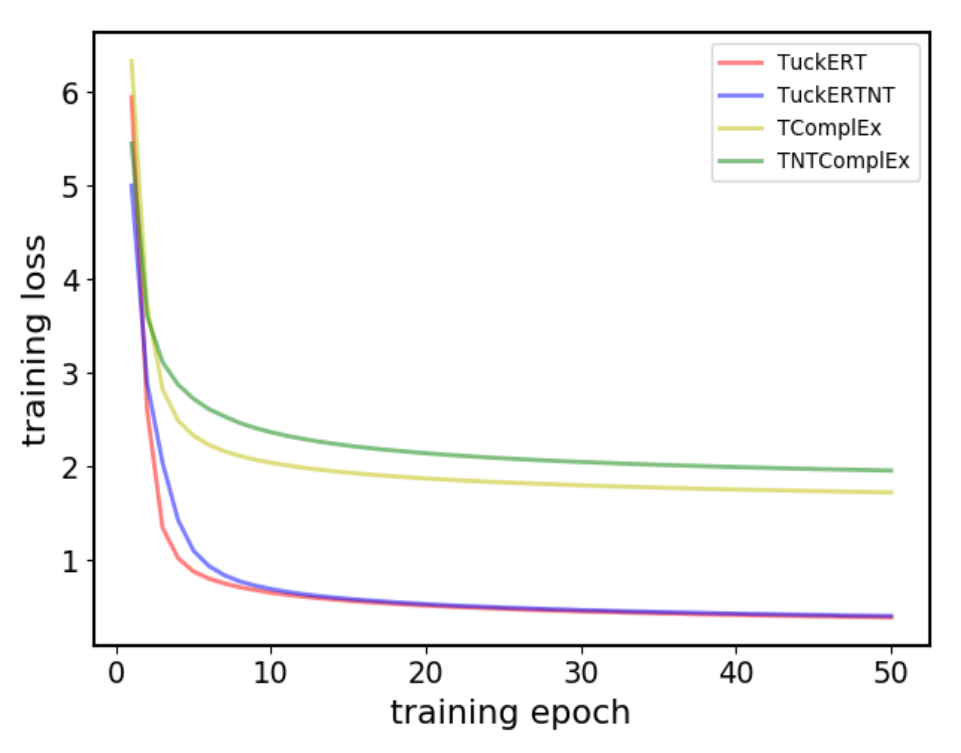} 
\caption{The training curve for the proposed models and baseline models.} 
\label{fig:: training_curve}
\end{figure}

To evaluate the stability of our model, Figure~\ref{fig:: training_curve} shows the curve of the training loss for the proposed models TuckERT, TuckERTNT, and baseline models TComplEx, TNTComplEx on the ICEWS14 dataset. According to the results of the curve comparison, we have the following two observations:
(1) On the ICEWS14 dataset, the TuckERT and TuckERTNT models have almost the same expressiveness in the training process, which suggests the two models may have close link prediction results on test sets. And the results in Table~\ref{tab:performance_comparison} also demonstrate this observation and assumption. (2) The training losses of the proposed models are lower than the baseline models TComplEx and TNTComplEx in the period of stabilization, it can be concluded that the proposed model is more expressive and robust.

\textbf{Discussion}

Theoretically, the time-wise information can be employed to associate with the subject, predicate or object to obtain time-dependent embedding equivalently as they express the same meaning. Here, we conduct experiments to verify whether the three models have the same performance. It is worth noting that binding the time information with object only can predict the subject or predicate, it is unfair to the first two models that predict the object, thus we only compare the performance of the first two models that binding time-wise information with the subject and predicate trained on the ICEWS14 and ICEWS2015. To evade the impact of the vector initialization, we present three experimental results of each model on two datasets. And the experimental results are presented in Table~\ref{tab:verifytimeassociate}. The results indicate that the performance of the two models is slightly different, irrespective of the factor of random initialization. The possible reasons for this are speculated as follows.

The performance of models that binding time with subject and predicate vary with different forms of data. Specifically, the model that binding time with predicate may has better performance on the data that reflects the temporal variation of the predicate, i.e. (Einstein, wasBornIn, Ulm, 1879), (Einstein, wonPrize, Nobel Prize, 1922), and (Einstein, diedIn, Princeton,1955). While the model that binding time with subject focuses on modeling the facts that have the same predicate, i.e. (Einstein, wasBornIn, Ulm, 1879), (Obama, wasBornIn, Hawaii, 1961), (Trump, wasBornIn, New York, 1946). The numerical values of the two models are equal in view of the associative law of multiplication, while they model different types of data. Which similar to the performance difference of predicting the head entity and tail entity on the 1-TO-Many and Many-TO-1 data. Therefore, the time-wise information can associate with the subject, predicate, and object equivalently in numerical representation, but they may have different performances as the dataset include different amounts of data in each type.
\renewcommand{\arraystretch}{1.8} 
\begin{table*}[tp]
\centering
\fontsize{7.5}{7}\selectfont
\begin{threeparttable}
\caption{The performace of models that associating time-wise information with the subject, predicate and object respectively trained on the ICEWS14 dataset.}
\label{tab:verifytimeassociate}
\begin{tabular}{l|llll|llll}
\toprule
\multirow{4}{*}{Datasets}&
\multicolumn{4}{c|}{$\left \langle \mathcal{W}; \mathrm{e}_s \odot \mathrm{e}_t , \mathrm{e}_r, \mathrm{e}_o\right \rangle$}&\multicolumn{4}{c}{ $ \left \langle \mathcal{W}; \mathrm{e}_s, \mathrm{e}_r \odot \mathrm{e}_t , \mathrm{e}_o\right \rangle$}\cr
\cmidrule(lr){2-5} \cmidrule(lr){6-9} 
&MRR&Hit@1&Hit@3&Hit@10&MRR&Hit@1&Hit@3&Hit@10\cr
\midrule
&0.540&0.466&0.578&0.681&0.548&0.470&0.588&0.695\cr
ICEWS14&0.538&0.463&0.576&0.680&0.547&0.468&0.587&0.694\cr
&0.540&0.466&0.576&0.680&0.547&0.469&0.588&0.695\cr
\hline
&0.563&0.484&0.605&0.709&0.571&0.495&0.610&0.716\cr
ICEWS2015&0.565&0.487&0.606&0.710&0.572&0.493&0.613&0.723\cr
&0.563&0.485&0.605&0.708&0.572&0.495&0.612&0.720\cr
\bottomrule
\end{tabular}
\end{threeparttable}
\end{table*}

\section{Conclusion and Outlook}
Tensor decomposition has been widely used in knowledge completion tasks, either on static KGs or temporal. In this work, we developed a new decomposition model that bridges the gap between tensor decomposition and temporal KGs completion from a generalized perspective. Based on this, we introduced several regularization schemes to study the impact of regularization on the proposed model. We proved that our method is fully expressive and achieves outstanding performance on three benchmarks compared with the existing state-of-the-art works. However, our models still include more parameters compared with previously presented models and have restrictions on efficiency when the embedding dimensionality exceeds the threshold. Future work might design a lightweight but powerful model, further work might consider that exploring a new paradigm for temporal KGs completion.

\section*{Acknowledgement}

\section*{References}
\bibliography{mybibfilespp}
\end{document}